\pdfoutput=1

\documentclass[11pt]{article}

\usepackage[]{ACL2023}

\usepackage{times}
\usepackage{latexsym}

\usepackage[T1]{fontenc}

\usepackage[utf8]{inputenc}

\usepackage{microtype}

\usepackage{inconsolata}
\usepackage{booktabs}
\usepackage{multirow}
\usepackage{xcolor} 
\usepackage{colortbl}
\definecolor{lightgray}{RGB}{219,226,233}

\title{Data and Approaches for German Text simplification – towards an Accessibility-enhanced Communication}

\author{Thorben Schomacker, Michael Gille, \\ \textbf{Jörg von der Hülls} \and \textbf{Marina Tropmann-Frick} \\
         Hamburg University of Applied Sciences \\ \texttt{thorben.schomacker@haw-hamburg.de} \\ \texttt{michael.gille@haw-hamburg.de}\\\texttt{joerg.vonderhuells@haw-hamburg.de} \\\texttt{marina.tropmann-frick@haw-hamburg.de}}

\begin{document}
\maketitle
\begin{abstract}
This paper examines the current state-of-the-art of German text simplification, focusing on parallel and monolingual German corpora. It reviews neural language models for simplifying German texts and assesses their suitability for legal texts and accessibility requirements.
Our findings highlight the need for additional training data and more appropriate approaches that consider the specific linguistic characteristics of German, as well as the importance of the needs and preferences of target groups with cognitive or language impairments.
The authors launched the interdisciplinary OPEN-LS \footnote{For more and up-to-date information, please visit our project homepage \url{https://open-ls.entavis.com}} project in April 2023 to address these research gaps. The project aims to develop a framework for text formats tailored to individuals with low literacy levels, integrate legal texts, and enhance comprehensibility for those with linguistic or cognitive impairments. It will also explore cost-effective ways to enhance the data with audience-specific illustrations using image-generating AI.

\end{abstract}

\section{Introduction}
In German-speaking countries, the majority of the population uses everyday language (Alltagssprache) in their daily affairs, with slight regional variations. However, in written texts, a more standardized 
vocabulary but with similar complexity \cite{bredelLeichteSpracheTheoretische2016} is typically preferred. In contrast, $12\%$ of the German population faces challenges in comprehending and utilizing standard language due to reduced literacy \cite{grotlueschenLEO2018Leben2020}. For more accessible and inclusive communication, this group depends on comprehensibility-enhanced language. Currently, specialized human translators convert standard language texts into simplified versions including easy language, with legal texts posing a particular challenge due to their technical nature and normative subject matter. 
Technical language texts represent one end of the complexity spectrum and easy language texts the other. This is further amplified by the fact that both text forms are linguistic expressions of 
constructed
languages.
To categorize training data effectively, we differentiate between "easy language" (Leichte Sprache) and "simple language" (einfache Sprache). "Easy language" refers to a highly comprehensible and rule-based form of German, whereby "simple language" is used to describe a variety of simplified language versions in the gray area between standard language and easy language \cite{maassEasyLanguagePlain2020}. Easy language is roughly equivalent with level A2 of the Common European Framework of Reference for Languages (CEFR).
Since public entities in Germany are required by law to translate information and communication texts into an accessible language version \cite{bgg11DisabilityEquality2022} the costs of this task burden the public budget. Automated approaches based on machine learning techniques promise to solve many of the challenges of text simplification, including the difficulties caused by technical language. A tool to simplify documents from different domains to a degree that facilitates these texts' comprehensibility for people with language or cognitive disabilities does not only improve understanding of these texts. It is also a key to inclusion and social participation \cite{unConventionRightsPersons2008}. This holds especially for domain-specific legal texts that are the starting point for the intra-lingual translation. 
In the course of our project 
we aim to build on existing simplification approaches using NLMs and adjust them with respect to the demands of the application domain. To achieve this objective, two specific aspects must be considered: First, the identification and systematic categorization of training data from the legal domain to build a quality-assured dataset to train a large language model for the domain specific simplification tasks in German. Second, the fine-tuning of an NLM under consideration of target-audience related comprehensibility requirements.
After a brief review of related work on German datasets and approaches (Section 2), this paper delivers a systematic assessment of published German datasets and approaches against the backdrop of the requirements of participatory communication of legal texts (Section 3). Finally, we outline the ongoing research project, "OPEN-LS: Open Data for Easy Language", which adopts a more target-group oriented approaches. We also identify and address several gaps in the existing research (Section 4). 
\section{Related Work}
Text simplification can be described as a machine translation task, converting one version of a language to another (Standard → Simple). However, compared to other machine translation tasks, automatic text simplification is a relatively new task. It started with a rule based statistical approach in 2010 \cite{speciaTranslatingComplexSimplified2010} on a small parallel Portuguese corpus (roughly 4,500 parallel sentences).  The first German simplification corpus was introduced in 2012 \cite{hanckeReadabilityClassificationGerman2012} and consisted of articles from GEO (similar to National Geographic) and GEOlino (GEO’s edition for children). In the initial paper, the corpus was only used for the training of statistical classifiers to predict the reading level of German texts. Their corpus was later improved and enlarged \cite{weissModelingReadabilityGerman2018}. In 2016 the first rule-based automatic text simplification system for German was released \cite{suterRulebasedAutomaticText2016}. In 2020 the first parallel corpus for data-driven automatic text simplification for German was published \cite{sauberliBenchmarkingDatadrivenAutomatic2020} and a first investigation of the use of a neural machine translation system for this problem in German was conducted. They concluded that the Austrian Press Agency corpus was not large enough to sufficiently train a neural machine translation system that produces both adequate and fluent text simplifications. In a later study, the same neural machine translation architecture was use and further evaluated concerning the levels of simplification which were generated by these models \cite{springExploringGermanMultiLevel2021}. In 2021 \citep{riosNewDatasetEfficient2021} adapted mBART \cite{liuMultilingualDenoisingPretraining2020} with Longformer Attention \cite{beltagyLongformerLongDocumentTransformer2020} and applied it to the task of document-level text simplification. It has been further explored on different domains, recently \citep{schomackerExploringAutomaticText2023,stoddenDEPLAINGermanParallel2023}. Furthermore, the first Decoder-only approach for German text simplification has been released \citep{anschutzLanguageModelsGerman2023}.

Automatic simplification of legal documents has only recently, in 2022, emerged \cite{collantesSimpaticoTextSimplification2015, cemriUnsupervisedSimplificationLegal2022, manorPlainEnglishSummarization2019, gallegosRightRemainPlain2022,gilleEinsatzNeuralLanguage2023,koppMachineTranslationEasy2023}. All of these works had to rely on monolingual datasets and state, that the task is still underinvestigated. To this day, there is no dataset with parallel legal documents (standard $\rightarrow$ simple language). In section 5.1 we will further discuss features and constraints of legal texts.
\section{Systematic and accessibility-oriented assessment of dataset landscape}
\subsection{Parallel Datasets}
To find all aligned German text simplification datasets, we focused our Google Scholar search on papers which prioritize German by including the word “German” in the title. Further, we wanted to find textual datasets, so used its synonyms: "dataset", "corpus", "data" or "texts". The task of text simplification can be covered by datasets with "simple" language or which investigate "readability" or text "complexity". So, we concluded on this query:
\texttt{allintitle: German corpus OR dataset OR data OR texts "Simple" OR "simplification" OR "readability" OR "complexity".}
This resulted in an identification of 14 parallel German datasets or sub-datasets as listed in Table \ref{tab:my_label}. By reading the dataset descriptions in their corresponding publication and checking the underlying data sources, we identified inductively text genres and domains in the dataset. We categorized them in three exclusive genres: 1) Encyclopedic (ENC) texts are summaries of knowledge either general or special to a particular field; 2) Articles (ART), are published nonfiction texts; and 3) Unknown (UNK), are texts, of which its author did not provide sufficient information to be clearly categorized. 
In addition to the genre, we tagged the datasets with seven domains: 1) Medical, which covers all aspects of human health; 2) Disability, which covers all aspects of the life and interests of people with disabilities; 3) News, are texts about current events without defining a field of interest; 4) Politics, discussing topics about politically viewpoints or activities such as electoral programs of political parties; 5) Government, any information, that is published by public authorities and/or containing administrative and non-partisan legal information; 6) Encyclopedic, collection of texts that could form a reference work without any specific field of interest; 7)  Unknown, are texts, of which its author did not provide sufficient information  to be clearly categorized. Aligned datasets thematically focused on legal aspects were not identified.
 We provide an overview of the datasets in Figure 1 by the number of documents. With a percentage of $73\%$ of the documents, News is the largest domain. The more practical and life-oriented categories Government, Disability and Medical are forming together less than $10\%$ of the available data. A significant proportion of $20\%$ of the available simple data is targeted to children. Training machine learning models with children-oriented simple language could lead to a bias. So, this type of data should be used with caution.

\begin{table*}[t!]
    \centering\fontsize{8pt}{8pt}\selectfont
{
    \begin{tabular}{l c c c c c c c}
    \bottomrule
       \rowcolor{lightgray} & Doc. & Simplicity &  &  & &  &  \\
       \rowcolor{lightgray} Name & Pairs & Versions & Genre & Domain & \multicolumn{2}{c}{Published} & URL \\
       \toprule
       20 Minuten & 18305 & STD, SIM & ART & News & \cite{riosNewDatasetEfficient2021} & 2021 & - \\
       KLEXIKON & 2899 & CH, AD & ENC & Encyclopedic & \cite{aumillerKlexikonGermanDataset2022} &2022 & \cite{aumillerKlexikonGermanDataset2023} \\
       APA & 2472 & A2, B1 & ART & News & \cite{sauberliBenchmarkingDatadrivenAutomatic2020} & 2021 & - \\
       (apo) & 2311 & STD, SIM & ART & Medical & \cite{toborekNewAlignedSimple2023} & 2022 & \cite{toborekNewAlignedSimple2022} \\
       Geo-Geolino & 1627 & CH, AD & ART & Science & \cite{hanckeReadabilityClassificationGerman2012} & 2022 & - \\
       Lexica & 1090 & CH, AD & ENC & Encyclopedic & \cite{hewettAutomaticallyEvaluatingConceptual2021} & 2021 & \cite{hewettLexicacorpus2022} \\
       capito & 752 & A1, A2, B1 & UNK & Unknown & \cite{riosNewDatasetEfficient2021} & 2021 & - \\
       Tagesschau / Logo & 415 & CH, AD & SUB & News & \cite{weissModelingReadabilityGerman2018} & 2018 & - \\
       & 378 & STD, SIM & ART & Unknown & \cite{battistiCorpusAutomaticReadability2020} & 2020 & - \\
       (bra), (mdr), (taz) & 377 & STD, SIM & ART & News & \cite{toborekNewAlignedSimple2022} & 2022 & \cite{toborekNewAlignedSimple2022} \\
       & 256 & CH, AD & ART & Disability & \cite{klaperBuildingGermanSimple2013} & 2013 & - \\
       (koe) & 82 & STD, SIM & ART & Government & \cite{toborekNewAlignedSimple2022} & 2022 & \cite{toborekNewAlignedSimple2022} \\
       (beb), (lmt) & 66 & STD, SIM & ART & Disability & \cite{toborekNewAlignedSimple2022} & 2022 & \cite{toborekNewAlignedSimple2022} \\
       TextComplexityDE & 23 & STD, SIM & ENC & Encyclopedic & \cite{seiffeSubjectiveTextComplexity2022} & 2019 & \cite{naderiTextComplexity2023} \\
       (soz) & 15 & STD, SIM & ART & Politics & \cite{toborekNewAlignedSimple2022} & 2022 & \cite{toborekNewAlignedSimple2022} \\
    \end{tabular}
   
    \caption{All available German parallel text simplification datasets and sub- datasets according to the Google Scholar results by using the query in section 1. For more details about the categorization, please refer to section 3.1. Simplicity Version are Standard Language (STD), any form of simple language (SIM), children-targeted (CH), adult-targeted-language (AD), and A1, A2, B1 are language level from the CEFR.  \label{tab:my_label}}
    }
    
\end{table*}

\subsection{Monolingual Datasets and Sources}
To gain a more complete picture of the datasets, we further investigated collections of German easy language, that have no standard language equivalent. Many newspapers or lexicons target children, e.g., ”Dein Spiegel” from ”Der Spiegel”. We decided to only include resources that use simple or easy language and did not research any children-targeted content because children-targeted content does not necessarily mean that it is accessible for the target we defined for simple and easy language.  Furthermore, we focused on resources that cover different genres to show the variety of genre currently used in easy language. Many text genres have no published parallel dataset despite the fact, that there are monolingual resources (e.g., narrative texts, legal texts). Similar to the parallel datasets, the majority of texts are news and encyclopedic articles. A comparatively large number of monolingual datasets address the interests of people with disabilities, not least because public authorities in Germany are obliged to communicate in simple and understandable language.

\subsection{(Non-)Consideration of Accessibility and Participation in Existing Datasets}
For the reasons outlined in Section 1 we focus on two particularly critical dimensions when considering accessibility and participation aspects for the evaluation of existing datasets and approaches: Legal texts and the concrete needs of the addressees.

\paragraph{Legal Texts:} Legal language comprises many different types of text such as laws and regulations, court judgements, witness statements, complaints, legal opinions etc. In addition, a large (and increasing) number of legal sub-domains, e.g., constitutional law, criminal law, AI law, exist. All of these different types of texts in the different (sub-)domains share similar linguistic traits, such as the use of legal jargon ('legalese'), formalization, long and complex sentences, a very high degree of intertextuality, mixed authorship (at least to some degree), a wide range of addressees and a unique tension between accuracy and vagueness \cite{baumannRechtstexteAlsBarrieren2020}. In addition, many legal texts are designed to be legally binding and establish rights and obligations. In establishing and organizing legal relationships these texts are fundamentally different from statements of fact that are subject of most intralingual and monolingual corpora. Thus, texts with legal content differ in many respects from texts in standard language. Furthermore, Legal texts fulfill certain text functions \cite{din-normenausschussergonomieEmpfehlungenFurDeutsche2023}. This text function, e.g. a legal binding, can deviate in the translation into plain language. These deviations should be consciously  handled. For these reasons alone, the training of neural language programs for the legal domain must be based on suitable German-language training materials.
\paragraph{Specialized format:} We pursue a participatory approach and collaborate with a large service provider and stakeholder of easy language recipients. The largest proportion of people with low literacy are disabled in some form. Most of them have difficulties to read texts, that exceed a half DIN A4 page, even if the text is written in easy language. Translating legal texts to a version that both maintains its meaning and is comprehensible to people with cognitive or language impairments, we need to define a specialized format. We propose the following four-level complexity hierarchy:
\begin{enumerate}
    \item A summary in easy language of the underlying standard language /legal document, which has a pre-defined maximum length. This text version should be easy to read and understand for people who need low barrier text forms. It also helps the reader to appreciate the central meaning of the underlying document.
    \item A longer version in easy language with jump markers that refer the reader to a glossary. This version is especially meant to be digital, so that the reader can access the glossary by a one-click action, that does not disturb the reading ﬂow.
    \item A complete version in easy language that should only reduce the linguistic complexity and not the complexity of content. It aims at conveying most of the (legal) statements of the original document. We assume that this version may be longer than the original text on which it is based.
    \item The original text in standard language.
\end{enumerate}

\section{Research gaps and planned contributions}
We identified and categorized existing resources for simplifying German texts with the aim of a preparatory assessment for the development of an NLM-based approach that supports accessible communication through participation-relevant texts. Our assessment of these intralingual-aligned and monolingual datasets as well as the existing approaches revealed the research gaps. Moreover, we observed that all monolingual datasets use illustrations to improve readability and intelligibility, while none of the parallel datasets do so. All identified datasets have a linear structure without any interactive elements, that could improve the readability. Based on our investigation and analyses in relation to the target group, we identify future areas of research:
\begin{enumerate}
    \item Identification and investigation of existing texts, which are tailored to the needs of the target group and improve the readability of texts both in monolingual and parallel datasets.
    \item Extension of parallel datasets by adding topics, domains and sub-domains, that are relevant for the everyday life of the target group.
    \item Addition of any form of illustration to the parallel datasets. By including visual elements, such as images, diagrams, or charts, the dataset becomes more inclusive and accessible to a wider range of users.
    \item The transferability of the model to domains and sub-domains (e.g., legal sub-domains) for which it has not been trained.
    \item The methodological development of evaluation methods that allow for an assessment that is in line with the objectives and purpose of accessibility, inclusion and participation by incorporating appropriate quantitative and qualitative methods. These evaluation methods may consider factors like readability scores, user feedback, comprehension tests, and other relevant metrics to measure the effectiveness of the model in promoting accessibility, inclusion, and participation.
\end{enumerate}

We want to tackle all five research gaps in the future, so that researchers and developers can enhance the quality and applicability of language models for the target group, making information more accessible and engaging for a broader audience. 
Our current focus is to make legal texts more accessible in German easy language. Documents from this domain are often pivotal to a self-empowered life. Based on texts in this domain, we aim at designing specialized accessibility-enhanced formats.

\newpage
\section*{Limitations}
In this work, we examined the current state-of-the-art of German text simplification. It reviews neural language models for simplifying German texts and assesses their suitability for legal texts and accessibility requirements. The general methodology of this paper is applicable for any domain or language, but only works for the task of text simplification. Furthermore, the review only focuses on German, so no definitive conclusions about the situation for other languages can be made based on this work alone. Additionally, this paper relied on the current draft version of the DIN standard \cite{din-normenausschussergonomieEmpfehlungenFurDeutsche2023}, the final version and its implications could deviate. Moreover, the DIN standard \cite{din-normenausschussergonomieEmpfehlungenFurDeutsche2023} is based on assumptions about its addressees, which we have not questioned further but simply adopted. These assumptions, e.g. include a homogeneity bias. Another limitation would be the limited use for pure information texts or transfers into information texts, i.e. that the target text function (in the sense of DIN) is always an informative one.

\section*{Ethics Statement}
This paper complies with the ACL Ethics Policy\footnote{\url{https://www.aclweb.org/portal/content/acl-code-ethics}}. The research field of this paper can help people to gain access to information by translating and transforming in an accessibility-enhanced way. Our presentation aims at motivating further scientific research and debate.

\section*{Acknowledgements}
This work is funded by UpdateHamburg (Schomacker), and by Hamburg Call for Transfer (von der Hülls). Furthermore, we thank the anonymous reviewers for their valuable feedback and discussion impulses.

\bibliography{acl2023}

\begin{thebibliography}{35}
\expandafter\ifx\csname natexlab\endcsname\relax\def\natexlab#1{#1}\fi

\bibitem[{Anschütz et~al.(2023)Anschütz, Oehms, Wimmer, Jezierski, and
  Groh}]{anschutzLanguageModelsGerman2023}
Miriam Anschütz, Joshua Oehms, Thomas Wimmer, Bart{\textbackslash}lomiej
  Jezierski, and Georg Groh. 2023.
\newblock \href {https://doi.org/10.18653/v1/2023.findings-acl.74} {Language
  {Models} for {German} {Text} {Simplification}: {Overcoming} {Parallel} {Data}
  {Scarcity} through {Style}-specific {Pre}-training}.
\newblock In \emph{Findings of the {Association} for {Computational}
  {Linguistics}: {ACL} 2023}, pages 1147--1158, Toronto, Canada. Association
  for Computational Linguistics.

\bibitem[{Aumiller(2023)}]{aumillerKlexikonGermanDataset2023}
Dennis Aumiller. 2023.
\newblock \href {https://github.com/dennlinger/klexikon} {Klexikon: {A}
  {German} {Dataset} for {Joint} {Summarization} and {Simplification}}.
\newblock Original-date: 2022-01-05T09:09:42Z.

\bibitem[{Aumiller and Gertz(2022)}]{aumillerKlexikonGermanDataset2022}
Dennis Aumiller and Michael Gertz. 2022.
\newblock Klexikon: {A} {German} {Dataset} for {Joint} {Summarization} and
  {Simplification}.
\newblock In \emph{Proceedings of the 13th {Conference} on {Language}
  {Resources} and {Evaluation} ({LREC} 2022)}, pages 2693--2701.

\bibitem[{Battisti et~al.(2020)Battisti, Pfütze, Säuberli, Kostrzewa, and
  Ebling}]{battistiCorpusAutomaticReadability2020}
Alessia Battisti, Dominik Pfütze, Andreas Säuberli, Marek Kostrzewa, and
  Sarah Ebling. 2020.
\newblock \href {https://aclanthology.org/2020.lrec-1.404} {A {Corpus} for
  {Automatic} {Readability} {Assessment} and {Text} {Simplification} of
  {German}}.
\newblock In \emph{Proceedings of the 12th {Language} {Resources} and
  {Evaluation} {Conference}}, pages 3302--3311, Marseille, France. European
  Language Resources Association.

\bibitem[{Baumann(2020)}]{baumannRechtstexteAlsBarrieren2020}
Antje Baumann. 2020.
\newblock \href
  {https://library.oapen.org/bitstream/handle/20.500.12657/43216/Handbuch%20Barrierefreie%20Kommunikation_OA.pdf?sequence=1#page=680}
  {Rechtstexte als {Barrieren} – {Einige} {Merkmale} der {Textsorte}
  "{Gesetz}" und die {Verständlichkeit}}.
\newblock In Christiane Maaß and Isabel Rink, editors, \emph{Handbuch
  {Barrierefreie} {Kommunikation}}, 1 edition, volume~3 of \emph{Kommunikation
  – {Partizipation} – {Inklusion}}, pages 679--702. Frank \& Timme, Berlin.

\bibitem[{Beltagy et~al.(2020)Beltagy, Peters, and
  Cohan}]{beltagyLongformerLongDocumentTransformer2020}
Iz~Beltagy, Matthew~E. Peters, and Arman Cohan. 2020.
\newblock \href {http://arxiv.org/abs/2004.05150} {Longformer: {The}
  {Long}-{Document} {Transformer}}.
\newblock ArXiv: 2004.05150.

\bibitem[{{BGG}(2022)}]{bgg11DisabilityEquality2022}
{BGG}. 2022.
\newblock \href {https://www.gesetze-im-internet.de/bgg/__11.html} {§ 11
  {Disability} {Equality} {Act} {BGG}}.

\bibitem[{Bredel and Maaß(2016)}]{bredelLeichteSpracheTheoretische2016}
Ursula Bredel and Christiane Maaß. 2016.
\newblock \emph{Leichte {Sprache} theoretische {Grundlagen}, {Orientierung}
  für die {Praxis}}.
\newblock Sprache im {Blick}. Dudenverlag.

\bibitem[{Cemri et~al.(2022)Cemri, Çukur, and
  Koç}]{cemriUnsupervisedSimplificationLegal2022}
Mert Cemri, Tolga Çukur, and Aykut Koç. 2022.
\newblock \href {http://arxiv.org/abs/2209.00557} {Unsupervised
  {Simplification} of {Legal} {Texts}}.
\newblock ArXiv:2209.00557 [cs].

\bibitem[{Collantes et~al.(2015)Collantes, Hipe, and
  Sorilla}]{collantesSimpaticoTextSimplification2015}
Miguel Collantes, Maureen Hipe, and Juan~Lorenzo Sorilla. 2015.
\newblock Simpatico: {A} {Text} {Simplification} {System} for {Senate} and
  {House} {Bills}.
\newblock In \emph{Proceedings of the 11th {National} {Natural} {Language}
  {Processing} {Research} {Symposium},}, volume~11, pages 26--32, Manila.

\bibitem[{{DIN-Normenausschuss
  Ergonomie}(2023)}]{din-normenausschussergonomieEmpfehlungenFurDeutsche2023}
{DIN-Normenausschuss Ergonomie}. 2023.
\newblock \href {https://doi.org/https://dx.doi.org/10.31030/3417293}
  {Empfehlungen für {Deutsche} {Leichte} {Sprache} ({DIN} {SPEC} 33429)}.

\bibitem[{Gallegos and George(2022)}]{gallegosRightRemainPlain2022}
Isabel Gallegos and Kaylee George. 2022.
\newblock The {Right} to {Remain} {Plain}: {Summarization} and {Simplification}
  of {Legal} {Documents}.

\bibitem[{Gille et~al.(2023)Gille, Schomacker, von~der Hülls, and
  Tropmann-Frick}]{gilleEinsatzNeuralLanguage2023}
Michael Gille, Thorben Schomacker, Jörg von~der Hülls, and Marina
  Tropmann-Frick. 2023.
\newblock \href {https://doi.org/https://doi.org/10.18420/rvi2023-013} {Der
  {Einsatz} von {Neural} {Language} {Models} für eine barrierefreie
  {Verwaltungskommunikation}: {Anforderungen} an die automatisierte
  {Vereinfachung} rechtlicher {Informationstexte}}.
\newblock In \emph{Proceedings of 6. {Fachtagung} {Rechts}- und
  {Verwaltungsinformatik} ({RVI} 2023)}, pages 144--158, Bonn. Gesellschaft
  für Informatik e.V.

\bibitem[{Grotlüschen and Buddeberg(2020)}]{grotlueschenLEO2018Leben2020}
Anke Grotlüschen and Klaus Buddeberg, editors. 2020.
\newblock \emph{{LEO} 2018: {Leben} mit geringer {Literalität}}.
\newblock wbv, Bielefeld.

\bibitem[{Hancke et~al.(2012)Hancke, Vajjala, and
  Meurers}]{hanckeReadabilityClassificationGerman2012}
Julia Hancke, Sowmya Vajjala, and Detmar Meurers. 2012.
\newblock \href {https://aclanthology.org/C12-1065} {Readability
  {Classification} for {German} using {Lexical}, {Syntactic}, and
  {Morphological} {Features}}.
\newblock In \emph{Proceedings of {COLING} 2012}, pages 1063--1080, Mumbai,
  India. The COLING 2012 Organizing Committee.

\bibitem[{Hansen-Schirra and Maaß(2020)}]{maassEasyLanguagePlain2020}
Silvia Hansen-Schirra and Christiane Maaß, editors. 2020.
\newblock \href {https://doi.org/10.26530/20.500.12657/42089} {\emph{Easy
  {Language} – {Plain} {Language} – {Easy} {Language} {Plus}: {Balancing}
  {Comprehensibility} and {Acceptability}}}, 1 edition, volume~3 of
  \emph{Easy–{Plain}–{Accessible}}.
\newblock Frank \& Timme, Berlin.
\newblock Accepted: 2020-09-28T09:51:54Z.

\bibitem[{Hewett(2022)}]{hewettLexicacorpus2022}
Freya Hewett. 2022.
\newblock \href {https://github.com/fhewett/lexica-corpus} {lexica-corpus}.
\newblock Original-date: 2021-08-13T09:12:24Z.

\bibitem[{Hewett and Stede(2021)}]{hewettAutomaticallyEvaluatingConceptual2021}
Freya Hewett and Manfred Stede. 2021.
\newblock \href {https://aclanthology.org/2021.konvens-1.23} {Automatically
  evaluating the conceptual complexity of {German} texts}.
\newblock In \emph{Proceedings of the 17th {Conference} on {Natural} {Language}
  {Processing} ({KONVENS} 2021)}, pages 228--234, Düsseldorf, Germany. KONVENS
  2021 Organizers.

\bibitem[{Klaper et~al.(2013)Klaper, Ebling, and
  Volk}]{klaperBuildingGermanSimple2013}
David Klaper, S.~Ebling, and Martin Volk. 2013.
\newblock \href {https://doi.org/10.5167/uzh-78610} {Building a
  {German}/{Simple} {German} {Parallel} {Corpus} for {Automatic} {Text}
  {Simplification}}.
\newblock In \emph{Klaper, {David}; {Ebling}, {S}; {Volk}, {Martin} (2013).
  {Building} a {German}/{Simple} {German} {Parallel} {Corpus} for {Automatic}
  {Text} {Simplification}. {In}: {The} {Second} {Workshop} on {Predicting} and
  {Improving} {Text} {Readability} for {Target} {Reader} {Populations} ({PITR}
  2013), {Sofia}, {Bulgaria}, 8 {August} 2013.}, pages 11--19, Sofia, Bulgaria.
  University of Zurich.

\bibitem[{Kopp et~al.(2023)Kopp, Rempel, Schmidt, and
  Spieẞ}]{koppMachineTranslationEasy2023}
Tobias Kopp, Amelie Rempel, Andreas~P. Schmidt, and Miriam Spieẞ. 2023.
\newblock \href {https://doi.org/10.57088/978-3-7329-9026-9_14} {Towards
  machine translation into {Easy} {Language} in public administrations:
  {Algorithmic} alignment suggestions for building a translation memory}.
\newblock In Silvana Deilen, Silvia Hansen-Schirra, Sergio~Hernández Garrido,
  Christiane Maaß, and Anke Tardel, editors, \emph{Emerging {Fields} in {Easy}
  {Language} and {Accessible} {Communication} {Research}}, volume~14, pages
  371--406. Frank \& Timme GmbH, Berlin.
\newblock Series Title: Easy – Plain – Accessible.

\bibitem[{Liu et~al.(2020)Liu, Gu, Goyal, Li, Edunov, Ghazvininejad, Lewis, and
  Zettlemoyer}]{liuMultilingualDenoisingPretraining2020}
Yinhan Liu, Jiatao Gu, Naman Goyal, Xian Li, Sergey Edunov, Marjan
  Ghazvininejad, Mike Lewis, and Luke Zettlemoyer. 2020.
\newblock \href {https://doi.org/10.1162/tacl\_a\_00343} {Multilingual
  {Denoising} {Pre}-training for {Neural} {Machine} {Translation}}.
\newblock \emph{Transactions of the Association for Computational Linguistics},
  8:726--742.
\newblock Place: Cambridge, MA Publisher: MIT Press.

\bibitem[{Manor and Li(2019)}]{manorPlainEnglishSummarization2019}
Laura Manor and Junyi~Jessy Li. 2019.
\newblock \href {https://doi.org/10.18653/v1/W19-2201} {Plain {English}
  {Summarization} of {Contracts}}.
\newblock In \emph{Proceedings of the {Natural} {Legal} {Language} {Processing}
  {Workshop} 2019}, pages 1--11, Minneapolis, Minnesota. Association for
  Computational Linguistics.

\bibitem[{Naderi(2023)}]{naderiTextComplexity2023}
Babak Naderi. 2023.
\newblock \href {https://github.com/babaknaderi/TextComplexityDE} {Text
  {Complexity} {DE}}.
\newblock Original-date: 2020-09-30T09:43:40Z.

\bibitem[{Rios et~al.(2021)Rios, Spring, Kew, Kostrzewa, Säuberli, Müller,
  and Ebling}]{riosNewDatasetEfficient2021}
Annette Rios, Nicolas Spring, Tannon Kew, Marek Kostrzewa, Andreas Säuberli,
  Mathias Müller, and Sarah Ebling. 2021.
\newblock \href {https://doi.org/10.18653/v1/2021.newsum-1.16} {A {New}
  {Dataset} and {Efficient} {Baselines} for {Document}-level {Text}
  {Simplification} in {German}}.
\newblock In \emph{Proceedings of the {Third} {Workshop} on {New} {Frontiers}
  in {Summarization}}, pages 152--161, Online and in Dominican Republic.
  Association for Computational Linguistics.
\newblock Tex.ids= riosNewDatasetEfficient2021a.

\bibitem[{Schomacker et~al.(2023)Schomacker, Dönicke, and
  Tropmann-Frick}]{schomackerExploringAutomaticText2023}
Thorben Schomacker, Tillmann Dönicke, and Marina Tropmann-Frick. 2023.
\newblock Exploring {Automatic} {Text} {Simplification} of {German} {Narrative}
  {Documents}.
\newblock In \emph{Proceedings of the 19th {Conference} on {Natural} {Language}
  {Processing} ({KONVENS} 2023)}.

\bibitem[{Seiffe et~al.(2022)Seiffe, Kallel, Möller, Naderi, and
  Roller}]{seiffeSubjectiveTextComplexity2022}
Laura Seiffe, Fares Kallel, Sebastian Möller, Babak Naderi, and Roland Roller.
  2022.
\newblock \href {https://aclanthology.org/2022.lrec-1.74} {Subjective {Text}
  {Complexity} {Assessment} for {German}}.
\newblock In \emph{Proceedings of the {Thirteenth} {Language} {Resources} and
  {Evaluation} {Conference}}, pages 707--714, Marseille, France. European
  Language Resources Association.

\bibitem[{Specia(2010)}]{speciaTranslatingComplexSimplified2010}
Lucia Specia. 2010.
\newblock \href {https://doi.org/10.1007/978-3-642-12320-7\_5} {Translating
  from {Complex} to {Simplified} {Sentences}}.
\newblock In \emph{Computational {Processing} of the {Portuguese} {Language}},
  Lecture {Notes} in {Computer} {Science}, pages 30--39, Berlin, Heidelberg.
  Springer.

\bibitem[{Spring et~al.(2021)Spring, Rios, and
  Ebling}]{springExploringGermanMultiLevel2021}
Nicolas Spring, Annette Rios, and Sarah Ebling. 2021.
\newblock \href {https://aclanthology.org/2021.ranlp-1.150} {Exploring {German}
  {Multi}-{Level} {Text} {Simplification}}.
\newblock In \emph{Proceedings of the {International} {Conference} on {Recent}
  {Advances} in {Natural} {Language} {Processing} ({RANLP} 2021)}, pages
  1339--1349, Held Online. INCOMA Ltd.

\bibitem[{Stodden et~al.(2023)Stodden, Momen, and
  Kallmeyer}]{stoddenDEPLAINGermanParallel2023}
Regina Stodden, Omar Momen, and Laura Kallmeyer. 2023.
\newblock \href {https://doi.org/10.48550/arXiv.2305.18939} {{DEPLAIN}: {A}
  {German} {Parallel} {Corpus} with {Intralingual} {Translations} into {Plain}
  {Language} for {Sentence} and {Document} {Simplification}}.
\newblock ArXiv:2305.18939 [cs].

\bibitem[{Suter et~al.(2016)Suter, Ebling, and
  Volk}]{suterRulebasedAutomaticText2016}
Julia Suter, Sarah Ebling, and Martin Volk. 2016.
\newblock Rule-based {Automatic} {Text} {Simpliﬁcation} for {German}.
\newblock In \emph{Proceedings of the 13th {Conference} on {Natural} {Language}
  {Processing}}, pages 279--287.

\bibitem[{Säuberli et~al.(2020)Säuberli, Ebling, and
  Volk}]{sauberliBenchmarkingDatadrivenAutomatic2020}
Andreas Säuberli, Sarah Ebling, and Martin Volk. 2020.
\newblock \href {https://aclanthology.org/2020.readi-1.7} {Benchmarking
  {Data}-driven {Automatic} {Text} {Simplification} for {German}}.
\newblock In \emph{Proceedings of the 1st {Workshop} on {Tools} and {Resources}
  to {Empower} {People} with {REAding} {DIfficulties} ({READI})}, pages 41--48,
  Marseille, France. European Language Resources Association.

\bibitem[{Toborek and Busch(2023)}]{toborekNewAlignedSimple2023}
Vanessa Toborek and Moritz Busch. 2023.
\newblock \href {https://github.com/mlai-bonn/Simple-German-Corpus} {A {New}
  {Aligned} {Simple} {German} {Corpus}}.
\newblock Original-date: 2022-08-22T10:58:53Z.

\bibitem[{Toborek et~al.(2022)Toborek, Busch, Boßert, Welke, and
  Bauckhage}]{toborekNewAlignedSimple2022}
Vanessa Toborek, Moritz Busch, Malte Boßert, Pascal Welke, and Christian
  Bauckhage. 2022.
\newblock \href {https://arxiv.org/abs/2209.01106} {A {New} {Aligned} {Simple}
  {German} {Corpus}}.

\bibitem[{{UN}(2008)}]{unConventionRightsPersons2008}
{UN}. 2008.
\newblock \href
  {https://www.ohchr.org/en/instruments-mechanisms/instruments/convention-rights-persons-disabilities}
  {{UN} {Convention} on the {Rights} of {Persons} with {Disabilities}
  ({CRPD})}.

\bibitem[{Weiß and Meurers(2018)}]{weissModelingReadabilityGerman2018}
Zarah Weiß and Detmar Meurers. 2018.
\newblock \href {https://aclanthology.org/C18-1026} {Modeling the {Readability}
  of {German} {Targeting} {Adults} and {Children}: {An} empirically broad
  analysis and its cross-corpus validation}.
\newblock In \emph{Proceedings of the 27th {International} {Conference} on
  {Computational} {Linguistics}}, pages 303--317, Santa Fe, New Mexico, USA.
  Association for Computational Linguistics.

\end{thebibliography}
\bibliographystyle{acl_natbib}

\end{document}